\documentclass[nonacm,sigconf]{acmart}

\usepackage{hyperref}
\usepackage{amsmath,amsfonts}
\usepackage{algorithmic}
\usepackage{graphicx}
\usepackage{textcomp}
\usepackage{xcolor}
\usepackage{soul}
\usepackage{booktabs}
\usepackage{multirow}
\usepackage{subcaption}

\newif\ifcomment
\commenttrue

\ifcomment
\newcommand{\stelios}[1]{\sethlcolor{stelios_colour}\hl{[\textbf{Stelios:} #1]}}
\newcommand{\steve}[1]{\sethlcolor{cyan}\hl{[Steve: #1]}}
\newcommand{\alex}[1]{\sethlcolor{orange}\hl{[Alex: #1]}}
\newcommand{\rui}[1]{\sethlcolor{green}\hl{[Rui: #1]}}
\newcommand{\nic}[1]{\sethlcolor{yellow}\hl{[Nic: #1]}}
\newcommand{\cut}[1]{\sethlcolor{light_red}\hl{[#1]}}
\else
\newcommand{\stelios}[1]{}
\newcommand{\steve}[1]{}
\newcommand{\alex}[1]{}
\newcommand{\rui}[1]{}
\newcommand{\nic}[1]{}
\newcommand{\cut}[1]{}
\fi

\usepackage{background}
\backgroundsetup{angle=0,
scale=1,
color=black,
firstpage=true,
position=current page.north,
hshift=0pt,
vshift=-20pt,
contents={\ifnum\value{page}=1 PREPRINT: Extended version of accepted paper at IEEE Pervasive Computing \else \fi}
}

\setcopyright{none}
\renewcommand\footnotetextcopyrightpermission[1]{} %
\begin{document}

\title{The Future of Consumer Edge-AI Computing}

\author{Stefanos Laskaridis}
\email{mail@stefanos.cc}
\affiliation{%
  \institution{Brave Software}
  \country{UK}
}

\author{Stylianos I. Venieris, Alexandros Kouris, Rui Li}
\email{{s.venieris,a.kouris,rui.li}@samsung.com}
\affiliation{%
  \institution{Samsung AI}
  \country{UK}
}

\author{Nicholas D. Lane}
\email{ndl32@cam.ac.uk}
\affiliation{%
  \institution{Flower Labs and \\University of Cambridge}
  \country{UK}
}

\renewcommand{\shortauthors}{Stefanos Laskaridis et al.}

\begin{abstract}

{In the last decade, Deep Learning has rapidly infiltrated the consumer end, mainly thanks to hardware acceleration across devices. However, as we look towards the future, it is evident that isolated hardware will be insufficient. Increasingly complex AI tasks demand shared resources, cross-device collaboration, and multiple data types, all without compromising user privacy or quality of experience. To address this, we introduce a novel paradigm centered around EdgeAI-Hub devices, designed to reorganise and optimise compute resources and data access at the consumer edge.
To this end, we lay a holistic foundation for the transition from on-device to Edge-AI serving systems in consumer environments, detailing their components, structure, challenges and opportunities.}

\end{abstract}

\settopmatter{printfolios=true}
\maketitle

\section{Introduction}
Since their very advent, Deep Neural Networks (DNNs) have been getting larger in their attempt to be more accurate without losing generality. Simultaneously, higher accuracies have also been a result of combining multiple models (ensembles or cascades) or inventing more exotic architectures,
manually
or automatically,
that offer higher capacity, better generalisation or fewer inductive biases~\cite{vaswani2017attention}.

More recently, there have been emerging trends in {Artificial Intelligence (AI)}, {generative or discriminative}, which are changing the computational landscape quite significantly. On the one hand, the training of hyperscale models~\cite{bommasani2021opportunities}
that act as foundations in latent spaces for solving a multitude of downstream tasks in one~\cite{gpt3}
or multiple~\cite{clip} modalities has been dominating computation in cloud AI. {Prominent examples include Large Language Models (LLMs), text-to-image generation (out-painting) or generative image composition (in-painting).}
On the other hand, as devices become more capable, an increasing number of DNNs are deployed on-device, oftentimes required to run simultaneously. Furthermore, the advent of fields like Federated Learning (FL) and personalisation introduce on-device training workloads.

\begin{figure}[t]
    \centering
    \includegraphics[width=.47\textwidth]{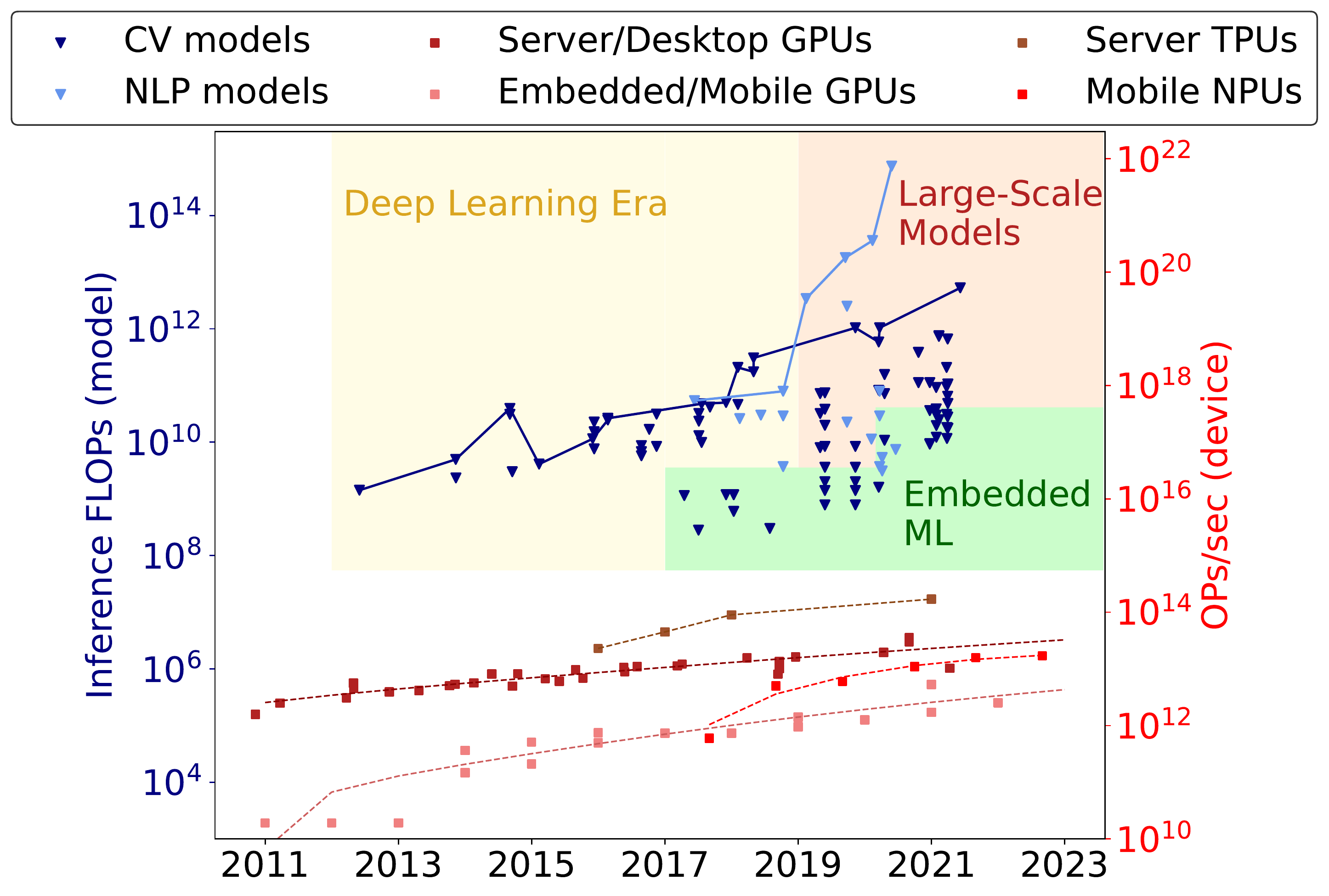}
    \caption{Evolution of DNNs operations (FLOPs) and hardware throughput (OP/s). Augmented data from \cite{desislavov2021compute}.} %
    \label{fig:ai-trajectory}
\end{figure}

Despite the forward-looking use-cases,
such workloads have been pushing the compute and memory requirements to unprecedented scales (Fig.~\ref{fig:ai-trajectory}), along with their data ingestion needs. {However, individual edge device capabilities have not scaled at the same pace.}
While the consumer edge becomes increasingly populated by smart devices, these continue to operate as standalone entities in isolation from their compute environment. Therefore, there are many missed opportunities for shaping a common context to learn and perform higher-level or fidelity tasks under a collaborative environment.

As such, a gap exists between compute requirements and resource availability for deploying intelligence at the consumer edge, {which is unlikely to be bridged only through traditional hardware scaling techniques.} %
In this paper, we present a new paradigm for organising resources at the {consumer} edge when executing emerging AI-tasks. Departing from isolated devices and moving into more capable EdgeAI-Hubs, we argue that the \textit{fluid sharing of compute} and the \textit{among-device sharing of context information} are key ingredients of an architecture that would deliver on the requirements of modern AI-tasks, with \textit{privacy} and \textit{sustainability} as vital components for deploying of state-of-the-art AI at the consumer edge.

\section{Deep Learning Trends}
\label{sec:ai-trajectory}

\begin{figure*}[t]
    \centering
    \includegraphics[width=.65\textwidth, trim={3.5cm 7cm 2.5cm 6cm}, clip]{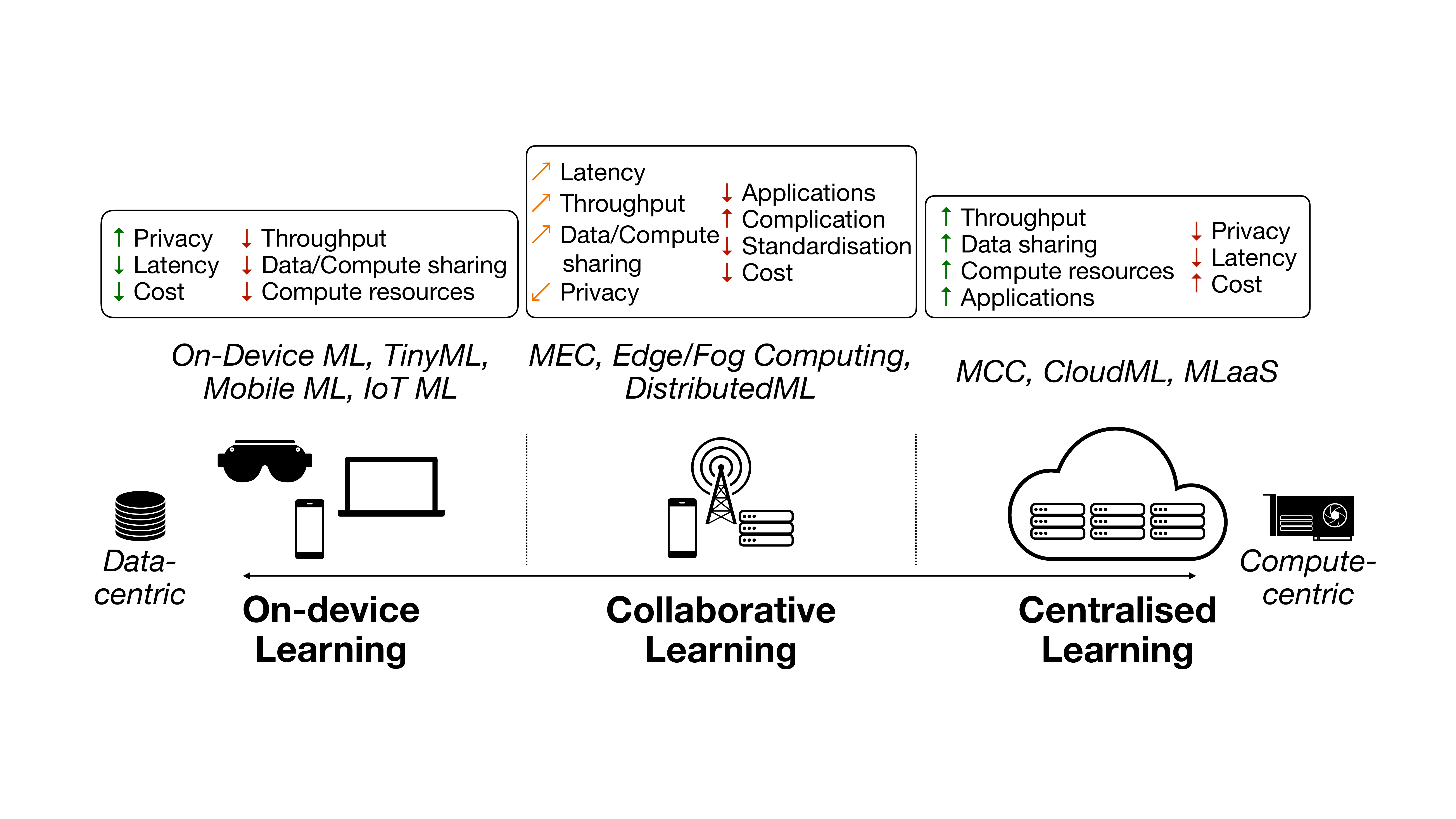}
    \caption{{Consumer Edge-AI 1.0 paradigms. While not necessarily mutually exclusive, they provide different levels of interaction between the involved entities.}}
    \label{fig:consumer-edgeai-1.0}
\end{figure*}

DNNs have traditionally grown in size in their strive for higher accuracies in pursuit of intelligence. 
Within a few years, we moved from the traditional multi-layer perceptron into deep networks of various forms and architectures. While the inductive bias of previous DNNs offered a convenient ``shortcut'' for learning in different modality inputs and a native mapping to widely available hardware, it turned out that convolutions were not ``all that we needed''.

The advent of transformers~\cite{vaswani2017attention} brought a breath of fresh air into deep learning, and enabled the training of hyperscale models, able to organise and query the knowledge over massive datasets, {and facilitated novel applications spanning across the personal and work life of the user. Such include NLP and vision related use-cases, such as next-gen multi-modal
chat assistants or meeting and document abstractive and creative tools.}
Unprecedented accuracies {and novel use-cases were} accompanied by inflated model footprints, only able to be trained distributedly in large datacenters. Even at inference, these models remain notoriously difficult to deploy, and much more so with real-time performance~\cite{laskaridis2024melting}. 
{Indicatively, running a 4-bit quantised version of Llama-2-7B on an M2 Max SoC (Metal) vs. a Galaxy S23 Ultra (OpenCL), yields 7.2$\times$ higher throughput, based on our experiments.}

Besides execution time, the large memory footprint of recent models has a direct energy impact on hardware, which becomes the main scaling bottleneck~\cite{Patterson2022-ix}. Memory accesses dominate energy, with more than 100$\times$ higher consumption than computation, while on-chip caches often account for 50\% of the processor's energy budget. %
Indicatively, executing 
TinyBERT (full-precision weights: 255MB)
on mobile processors, such as the Edge TPU (8MB cache),
requires an excessive amount of off-chip memory accesses and intensive use of the on-chip cache~\cite{Patterson2022-ix}.
As a result, even if new mobile processors could execute such models faster, the battery would be drained at an unacceptable rate.

At the same time, privacy of user data has become a top priority. %
Thus, alternative models of decentralised training have appeared, namely FL and on-device personalisation. While game-changing, since it enables collaborative learning and pushes computation to where data reside, training on device creates a much more memory- and energy-consuming workload that not all mobile devices can support. Indicatively, training SmallBERT on device can consume more that 8GB of peak memory,
while inference requires 1/16th of that~\cite{memory-eff-training-on-device}. Simultaneously, such in-the-wild deployments need to deal with data and system heterogeneity, along with partial availability and dubious robustness of clients, {while competing for resources with other co-habitating workloads.}

\section{The Evolution of ML Compute}
\label{sec:hw-trajectory}

The adoption and continuous upscaling of DNNs would not have been possible without the scaling of compute capabilities. %
GPUs, and their re-purposing from graphical to neural accelerators, initially paved the way towards training deep networks, followed by specialised ASIC/FPGA \mbox{accelerators.}

\noindent
\textbf{On-device compute for DNNs.}
Nevertheless, DNNs were never intended to stay limited to the premises of the datacenter.
AI advancements made their way into the consumer world~\cite{gaugenn2021imc} with smart devices of different shapes and forms, ranging from smartphones and wearables to IoT devices and even robots. Their omnipresence {and sensing capabilities} enabled the production of unprecedented scales of local data from various modalities, available for harvesting.

While initially fully offloaded to the
cloud at the cost of privacy and latency, DNN computation has since progressively been ``onloaded'' to devices.
What made this possible was advances in mobile hardware, which leverage integrated GPUs and Neural Processing Units (NPUs) on the same System-on-Chip (SoC)~\cite{samsung_npu} or external accelerators, \textit{e.g.}~Edge TPU.
In addition, EfficientML techniques have enabled performant on-device execution for a multitude of AI-tasks.

Still, in contrast to typical cloud-based infrastructure, the computational landscape in-the-wild is much more \textit{heterogeneous}~\cite{embench2019emdl,gaugenn2021imc,laskaridis2024melting}
and \textit{failure-prone}~\cite{fjord,fl_sys_design2019mlsys},
while the device still needs to remain responsive for tasks of the respective user. This brings new challenges that need to be tackled by device manufacturers and developers for making the mobile devices truly ``smart.''

\noindent
\textbf{On-device hardware vs. offloading.}
DNN compute requirements and the respective hardware have not necessarily scaled at the same pace {(Fig.~\ref{fig:ai-trajectory})}, as their development cycles are very different. In essence, developing and deploying highly specialised hardware is a high-cost process and needs to be balanced from a utility-cost standpoint
and from a device lifecycle perspective.
Regardless, cloud infrastructure still remains much more capable and allows for performant general-purpose compute without the constraints or heterogeneity of the edge. {However, this occurs at the expense of client latency, provider cost and privacy.} Thus, the ubiquitous question on  how much computation to offload or whether to dedicate on-device hardware is posed in the consumer \mbox{Edge-AI space.} %

{\section{The State of Consumer Edge-AI}}
\begin{table*}[t]
    \caption{{Consumer Edge-AI 2.0 enabling technologies, materialising the pillars of the proposed paradigm. They can be leveraged to achieve sustainable performance gains, preserve privacy and enable cross-device collaboration.}}
    \label{tab:related-tech}
    \resizebox{\linewidth}{!}{
        \begin{tabular}{@{} lll @{}}
        \toprule
        \textbf{Technology} & \textbf{Insights} & \textbf{Related Work} \\
        \midrule
        \textbf{Shared Compute} \\
        \quad Multi-DNN execution \& model serving & Scheduling multiple DNNs on embedded hardware & \cite{venieris2022multi} \\
        \quad Offloading/Split learning & Partially offload DNN computation to faster remote endpoint & \cite{spinn} \\
        \quad Hardware/DNN co-design & Hardware and DNN architecture co-design for tighter intergration & \cite{codesign} \\
        \quad Multi-radio communications & Leverage and load-balance between multiple channels in a priority-aware manner  & \cite{xie2015interference} \\
        \textbf{Shared Context} \\
        \quad Multi-view classification & Use of multiple sensors to collaboratively complete the same task & \cite{multiview} \\
        \quad Multi-task learning & Shared DNN architecture to target multiple downstream tasks & \cite{smith2017federated} \\
        \quad Collaborative learning & Confidential and private collaborative learning between agents  & \cite{capc2021iclr} \\
        \textbf{Privacy} \\
        \quad Trusted Execution Environments (TEEs) & Secure enclaves for attested and verifiable execution & \cite{darknetz} \\
        \quad Differential Privacy (DP) & Locally or globally added noise for privacy preservation &  \cite{mcmahan2017learning} \\
        \quad Secure Aggregation (SecAgg) & Aggregation based on Secure Multi-party computation primitives & \cite{bonawitz2019federated} \\
        \quad Homomorphic Encryption (HE) & Compute based on encrypted data &
        \cite{fl_homoencryption_usenix} \\
        \textbf{Sustainability} \\
        \quad Model Compression \& Efficient ML & Techniques to leverage model redundancy and extraneous precision for compression & \cite{wang2019deep,wan2024efficient} \\
        \quad Early-exiting & Intermediate exits along the DNN to preemptively stop computation on easier examples & \cite{10.1145/3469116.3470012,laskaridis2020hapi} \\
        \quad Heterogenous Multi-Processors \& DVFS & H/W with heterogeneous cores and Dynamic Voltage and Frequency Scaling for energy savings & \cite{10.1145/3470974} \\
        \quad Application-specific benchmarking \& optimisations & Need for per-application cross-hardware benchmarks and optimisations & \cite{jiang2018chameleon,9001257} \\

        \bottomrule
        \end{tabular}
    }
\end{table*}

Currently, there are emerging
applications in the consumer domain, still too costly to be deployed to the edge, or lack a common context that could be shared amongst devices in the ecosystem. In this setting, {referred as Consumer Edge-AI 1.0}, there is a lot of unfulfilled potential that is currently limited by the way intelligence is organised, deployed and distributed.

\subsection{Devices are islands}

While more and more devices penetrate the realm of the smart edge, they generally remain siloed, integrating standalone hardware to support their narrow requirements and {generally} do not advertise their capabilities to other local devices. {Moreover, separate personal and work devices tend to be a common setup for legal and privacy reasons.} In such an inelastic deployment, {not only are many compute cycles wasted}, but devices remain constrained in their own settings, and replicate similar sensor integration to perform different tasks.  As such, the per-device cost is higher, the utilisation remains low and the hardware investment gets retired at the end of the device's lifecycle. {While computation \textit{offloading} and \textit{split} computing~\cite{spinn} have been widely proposed before, they remain largely point-to-point and lack {horizontal} sharing of contextual information amongst user devices, therefore {acting} as mere remote accelerators. Thus, for personalising one's own GPT-assistant, individual devices would need to reach and offload data to third-party cloud services, as they currently lack a common data and compute fabric.}

\subsection{Shortcomings of previous paradigms}

There have been various paradigms of organising ML execution between devices, offering varying levels of success and adoption, {presented in Fig.~\ref{fig:consumer-edgeai-1.0}.}
Specifically:

\noindent\textbf{On-device ML~\cite{9586232}.}
The one end of the spectrum is to run AI-tasks locally on-device. While simple, it is far from simplistic as embedded and mobile devices come with severe constraints across computation, energy and thermals~\cite{laskaridis2024melting}. Thus, specialised hardware integration along with EfficientML techniques become core enablers~\cite{gaugenn2021imc}. {Data-sharing remains minimal.}

\noindent{\textbf{Centralised Learning}.
The other end suggests that storage and compute are outsourced from the mobile device to more powerful cloud resources. Initially theorised through the Mobile Cloud Computing (MCC)~\cite{yousefpour2019all} paradigm, it remains today the status-quo method for training and deployment of ML workloads. Latency and privacy remain key issues of this paradigm, along with the need to transfer all data.}

\noindent{\textbf{Collaborative Learning.}
Edge computing~\cite{yousefpour2019all} has been motivated by the movement of compute closer to where the data are produced, \textit{i.e.}~the edge. It provides lower latency compared to cloud offloading (\textit{i.e.}~MCC) and can be anywhere in between the cloud and the end-device, including base station servers (\textit{i.e.}~MEC) and ambient devices (\textit{i.e.}~Fog Computing). These paradigms mainly focused on general compute sharing, and while forward-looking, they never found a ``killer'' use-case driving their adoption, and devices remained isolated with opportunistic collaboration if any.
Closer to ML, this paradigm can manifest as Distributed Learning, with use-cases including inference offloading and distributed training.}

\section{Consumer Edge-AI 2.0}
\label{sec:edge-ai}

Given the accelerated rise of new AI-based use-cases and the inability of {individual} smart devices to scale up their capabilities {and context} to that dynamic, we propose a new architecture that aims at ML execution, through flexible sharing of compute resources with smarter placement of specialised hardware, so that edge devices' processing power is augmented {sustainably} without extreme duplication.
At the same time {and contrary to prior paradigms}, their sensing capabilities can also be enhanced via sharing of their situational context {to enable collaboration while respecting privacy}.
{We erect our proposed architecture around the following pillars, further supported by techniques presented in Tab.~\ref{tab:related-tech}.}

\subsection{Shared compute}

\noindent
\textbf{Hardware resource allocation.} Specifically, we shift from the paradigm of developing and integrating highly specialised NPUs into every available device to a model where higher-end and more general accelerators can be hosted in \textit{central-device hubs}. These hubs can be standalone devices that only serve this purpose (\textit{e.g.}~a home accelerator) or ``piggy-backed'' into devices with longer life cycles omnipresent at the edge (\textit{e.g.}~a router or TV). Thus, hardware can take advantage of larger budget in terms of \textit{area}, \textit{cooling} and \textit{power} to provide acceleration to more operations and users than before.
In fact, this hardware can even be based on reconfigurable FPGAs~\cite{venieris2022multi},
so that they can be adjusted to the user needs and available devices.
Last, they can be designed in a way that many hubs can be interconnected to scale out capabilities. %

We distinguish two levels of compute sharing, {both of which act complimentarily to one another and need to be optimised in tandem to achieve resource allocation efficiency}: \textit{i)}~\textit{static partitioning} and \textit{ii)}~\textit{dynamic resource sharing}. \textit{Static partitioning} refers to the placement of dedicated compute units to devices. While certain IoT devices can become as thin as a sensor with a network adapter, others still need to maintain some autonomy or mobility. As such, one cannot simply centralise all available compute and strip devices out of any meaningful capabilities. {Instead}, a balance between the two ends of the spectrum is needed.
For example, on-device inference may be a common AI-task, but opportunistic participation\footnote{Participation in FL usually happens when device is plugged in and connected to WiFi.} in FL has not shown a real need so far. Thus, smartphones may not need train-capable hardware integration. Instead, a training-ready NPU could be integrated to a home hub where training can be offloaded.

\begin{figure}[t]
    \centering
    \includegraphics[trim={0.5cm 16cm 0.5cm 0.9cm},clip,width=.42\textwidth]{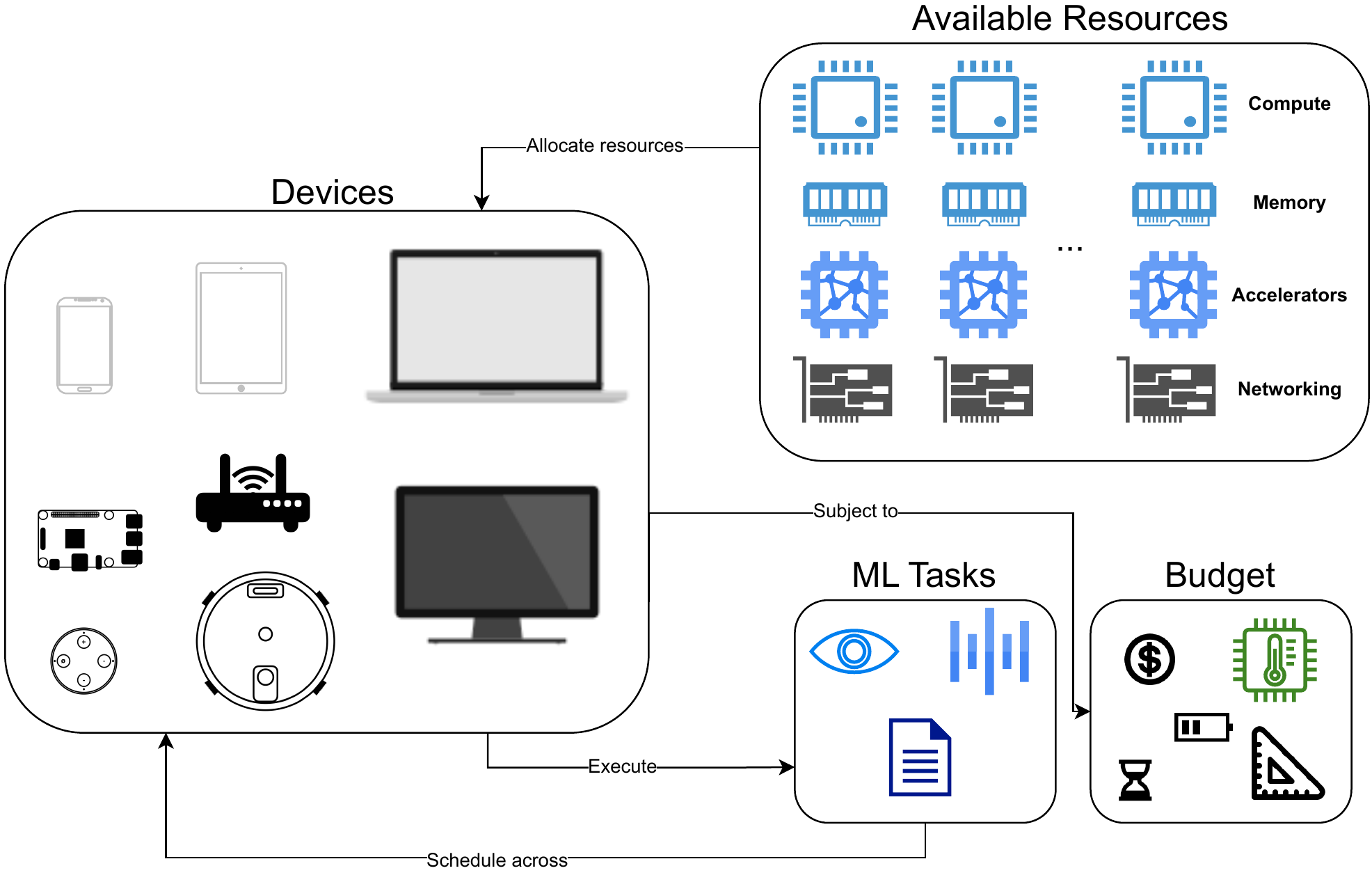}
    \vspace{-0.05cm}
    \caption{Resource allocation for the execution of ML tasks under a specific budget.}
    \label{fig:knapsack_resources}
\end{figure}

Simultaneously, \textit{resource sharing} can happen \textit{dynamically} between device-hub or in a \textit{peer-to-peer} manner, compatibly with {Edge/}Fog Computing~\cite{yousefpour2019all}. Workloads can be offloaded on demand, based on either the static capabilities of the device or their instantaneous dynamic load. {Effectively, one can think of resource partitioning and allocation as a generalised Knapsack Problem, as depicted in Fig.~\ref{fig:knapsack_resources}.}

How the equilibrium of resource partitioning is met
is a function of \textit{i)}~the \textit{nature} of AI-tasks at hand, \textit{ii)}~their \textit{urgency} to complete, and \textit{iii)}~the \textit{intersection} of tasks between devices.

\noindent
\textbf{Networking \& scheduling.} Central to this elastic compute is a capable \textit{multi-channel networking} infrastructure on top of which different requests for AI-tasks can be \textit{scheduled} and \textit{prioritised}. Such an infrastructure should offer multi-channel access, as different devices support diverse protocols (\textit{e.g.}~Wi-Fi, BLE, Zigbee, LoRa, etc.), load-balancing among available channels and bandwidth slicing for better {Quality of Experience (QoE).} {Advancements in mobile wireless communications (B5G, 6G) can also facilitate the adoption of this paradigm, over sub-THz, THz or visible light communications (VLC)~\cite{dang2020should}.}
Additionally, task deadlines with preemption under multi-tenancy are core features for the scheduler to guarantee QoE for all active users.
For example, the upscaling of live streaming video for a user would need to be given higher priority than the classification of newly acquired gallery photos on a user's device, which can be done offline. {Thus, it is one of the hub's primary objectives not only to provide resources, but also to co-ordinate them.}

\subsection{Shared context}

By having interconnected compute, it is also possible to share a common context between sensing devices to fulfill collaborative tasks.
As such, a smart speaker may not only serve as a standalone device, but could also be used for recognising a user in a room {along with their intent (e.g.~work or entertainment)}
to personalise their experience by preloading their profiles on ambient devices. At the same time, it can serve as a secondary microphone for noise cancelling or even for intrusion detection along with a smart camera.
Context sharing can be \textit{i)}~\textit{explicit}, through sensor-data exchange or \textit{ii)}~\textit{implicit}, by embedding subsets of available sensors into a common subspace that can be leveraged for different tasks.
Last, different tasks can also share common DNN backends, instead of replicating them per device. For example, the obstacle detection subsystem of a robotic vacuum cleaner could share parts of a DNN with a pet surveillance camera.
In fact, both can act as sensors for enhancing the classification result through multi-view perspectives, with one even offering active vision capabilities.

\begin{figure}[t]
    \centering
    \includegraphics[width=.49\textwidth,trim={0 0 0 1cm},clip]{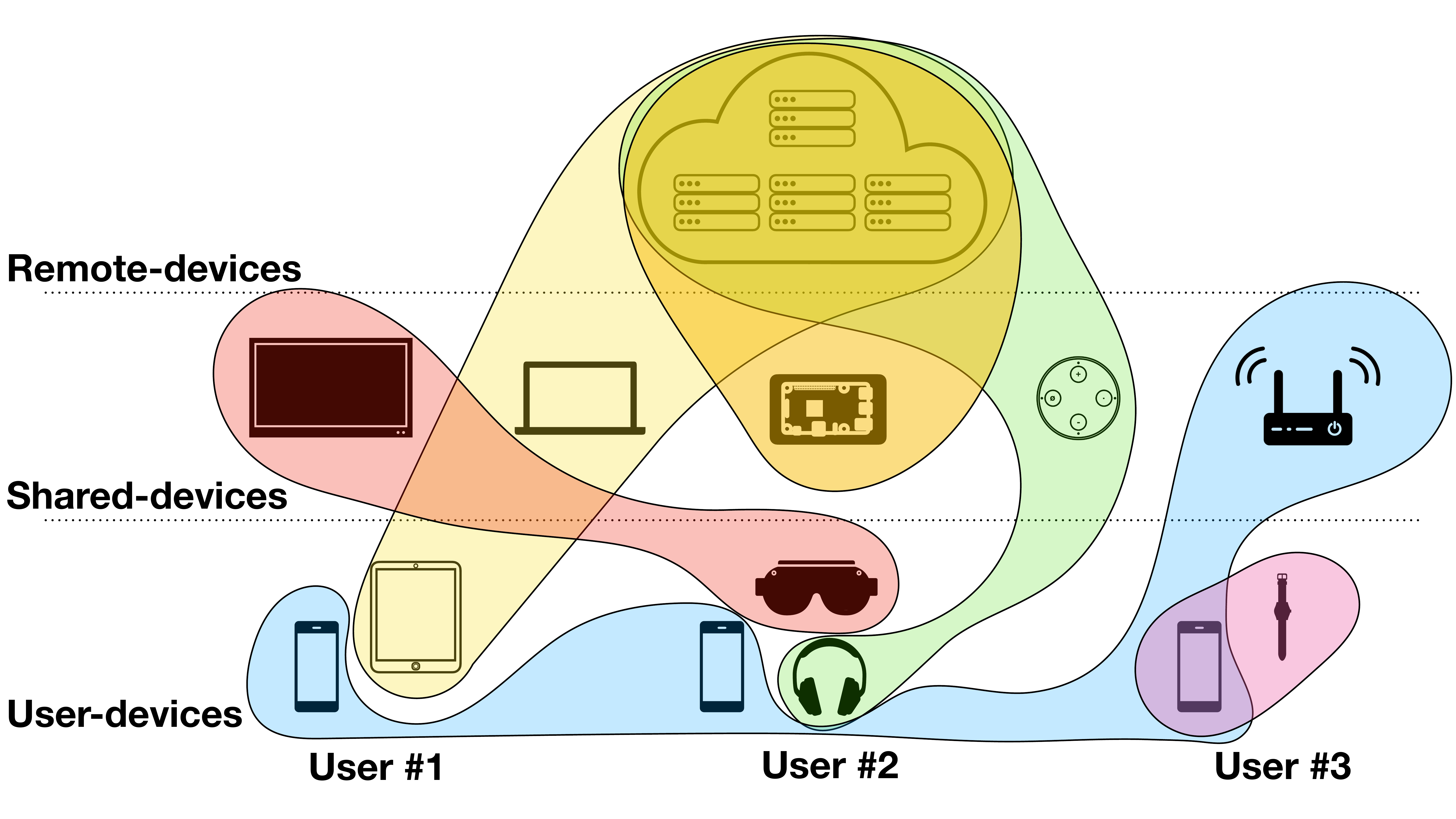}
    \caption{Privacy and collaboration zones between devices. {These can be supported via rights management through device-owner groups and Access-Control Lists.}}
    \label{fig:edgeai}
\end{figure}

\subsection{Privacy}

In this era of AI{, where information is eagerly digested by billion-scale models,} consumer data are increasingly important to protect.
Therefore, privacy becomes an innate design component of next-generation consumer Edge-AI.

\noindent\textbf{Privacy-preserving ML.}
Specifically, we differentiate between privacy and robustness at inference and training time. In the former setting, the input and outputs\footnote{Along with the intermediate representations.} of the DNN are sensitive, rather than the model itself.\footnote{From the user's standpoint. Model privacy may be of business-critical importance to the provider.} However, when training or personalising a model, along with the data involved, the model parameters and gradients become also privacy-sensitive, as they may leak user data. Overall, privacy
spans in a continuous spectrum, \textit{i.e.}~a privacy budget~\cite{mcmahan2017learning}, and therefore should be treated accordingly.

\noindent\textbf{Trust zones.} Simultaneously, different data have different sensitivity towards different entities/adversaries. For example, there are data that one might want accessible from home-owned devices but private to the public, such as holiday trip photos. Conversely, one could share their browsing preferences with a third-party for ad personalisation, but not with their household.
{The same applies between work and personal devices, a delineation that might not be straightforward in working-from-home settings.} As such, trust zones are formed that shape the data flow of collaboration in Edge-AI, depicted in Fig.~\ref{fig:edgeai}.

\subsection{Sustainable-AI}

Additionally to user privacy, it is also important to respect the environmental ecosystem in which Edge-AI operates. {We capitalise on the fact that not all downstream tasks require overprovisioned DNNs running round the clock and save on energy and {transmission} costs.} Simultaneously, we acknowledge the downsides of onloading more computation to the consumer edge where energy sources and efficiency may be suboptimal compared to large datacenters~\cite{Patterson2022-ix,Wu_undated-ns} {and propose ways to offset them.}
\cite{Wu_undated-ns}

\noindent\textbf{EfficientML.} A particularly important component in any embedded deployment of AI is the optimisation of resources. Embedded and mobile devices integrate lower computational capabilities than state-of-the-art servers.
To that direction, techniques from EfficientML~\cite{9586232} reduce the computational, memory or bandwidth requirements of AI-tasks, by means of changing the \textit{architecture},
\textit{representation} or \textit{execution}~\cite{fjord,10.1145/3469116.3470012} of the DNN in an offline or dynamic manner. {These, not only affect the users' QoE, but can drastically reduce energy consumption.} {For example, early-exit networks can leverage the difficulty~\cite{laskaridis2020hapi}
or spatio-temporal relatedness~\cite{kouris2022multi} of inputs to preempt computation on-demand.}

\noindent\textbf{Lower manufacturing costs.} Our paradigm also economises on the placement of highly-specific accelerators. By centralising certain components of the edge compute infrastructure, the manufacturing cost of mobile or IoT SoCs can be lower and the effective utilisation rate higher.

\noindent\textbf{Device upcycling.} The unprecedented pace of computer systems advancement means that devices become quickly deprecated and turn from useful companions to effectively \mbox{e-waste}.
However, old devices still integrate various sensors and oftentimes enough compute power to be useful~\cite{203326}. We embrace the previous-generation devices co-existence and integrate them in a sustainable manner to the smart edge context, through upcycling and repurposing.

\section{Proposed Architecture}
\label{sec:proposed_arch}

Here, we lay out a reference architecture of the systems implementing our paradigm and how different components interact to accomplish the end goal, \textit{i.e.}~supporting and enabling AI-task execution through a common computational and contextual fabric, while preserving user-privacy and sustainability. {We select an edge-accelerated solution as on-device resources are clearly not sufficient for the upcoming workloads (Fig.~\ref{fig:ai-trajectory}) and \textit{fully} offloading to resources beyond the edge hurts user-privacy and racks up provider costs.}

\begin{figure*}[t]

    \captionsetup[subfigure]{labelformat=empty}
    \centering
    \subfloat[]{
        \centering
        \vspace{0.1cm}
        \includegraphics[width=1.\columnwidth,trim={0cm -2cm 0cm 0cm},clip]{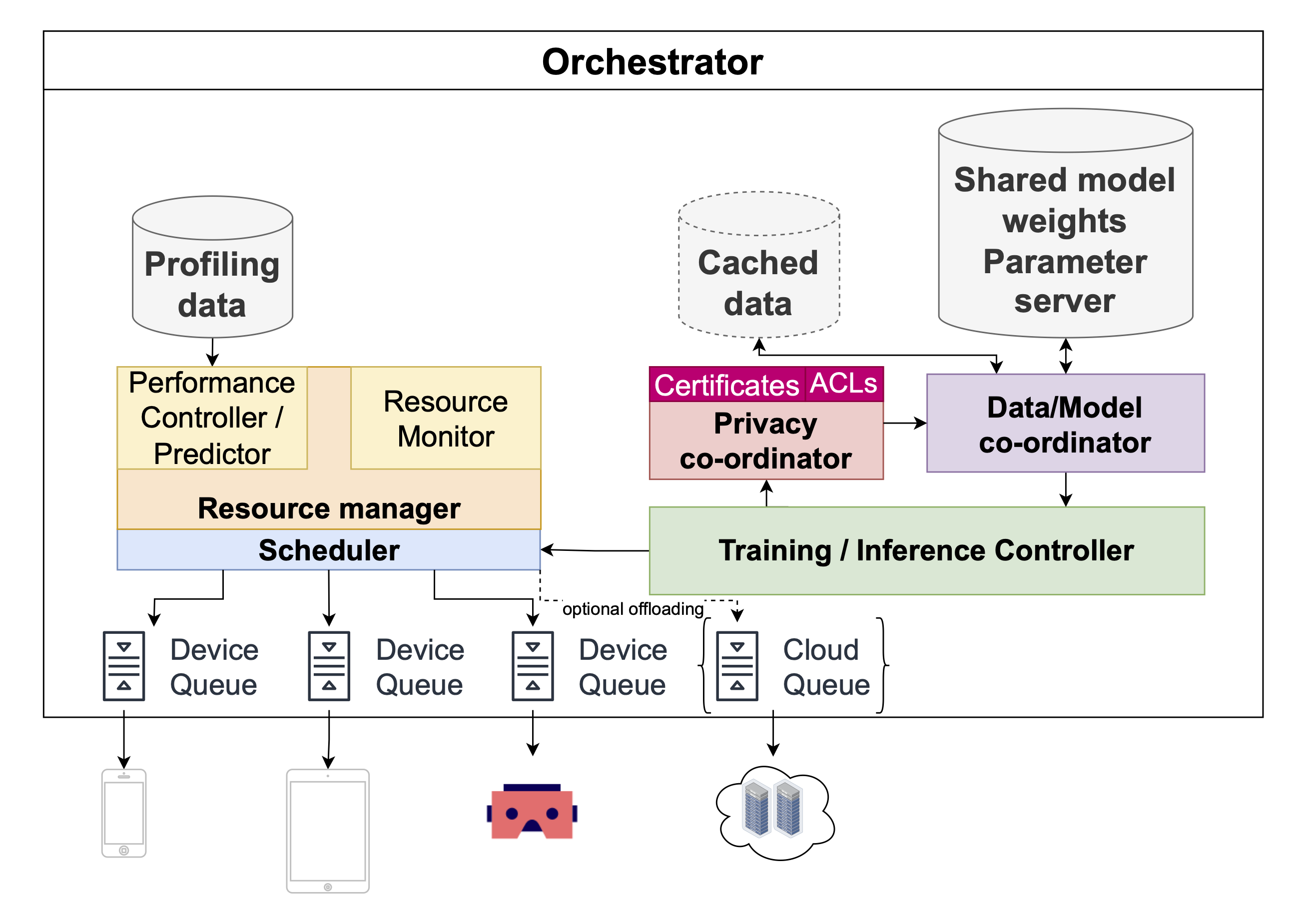}
        \label{fig:edgeai_ref}
    }
    \subfloat[]{
        \centering
        \includegraphics[width=1.01\columnwidth,trim={0cm -13cm 0cm 0cm},clip]{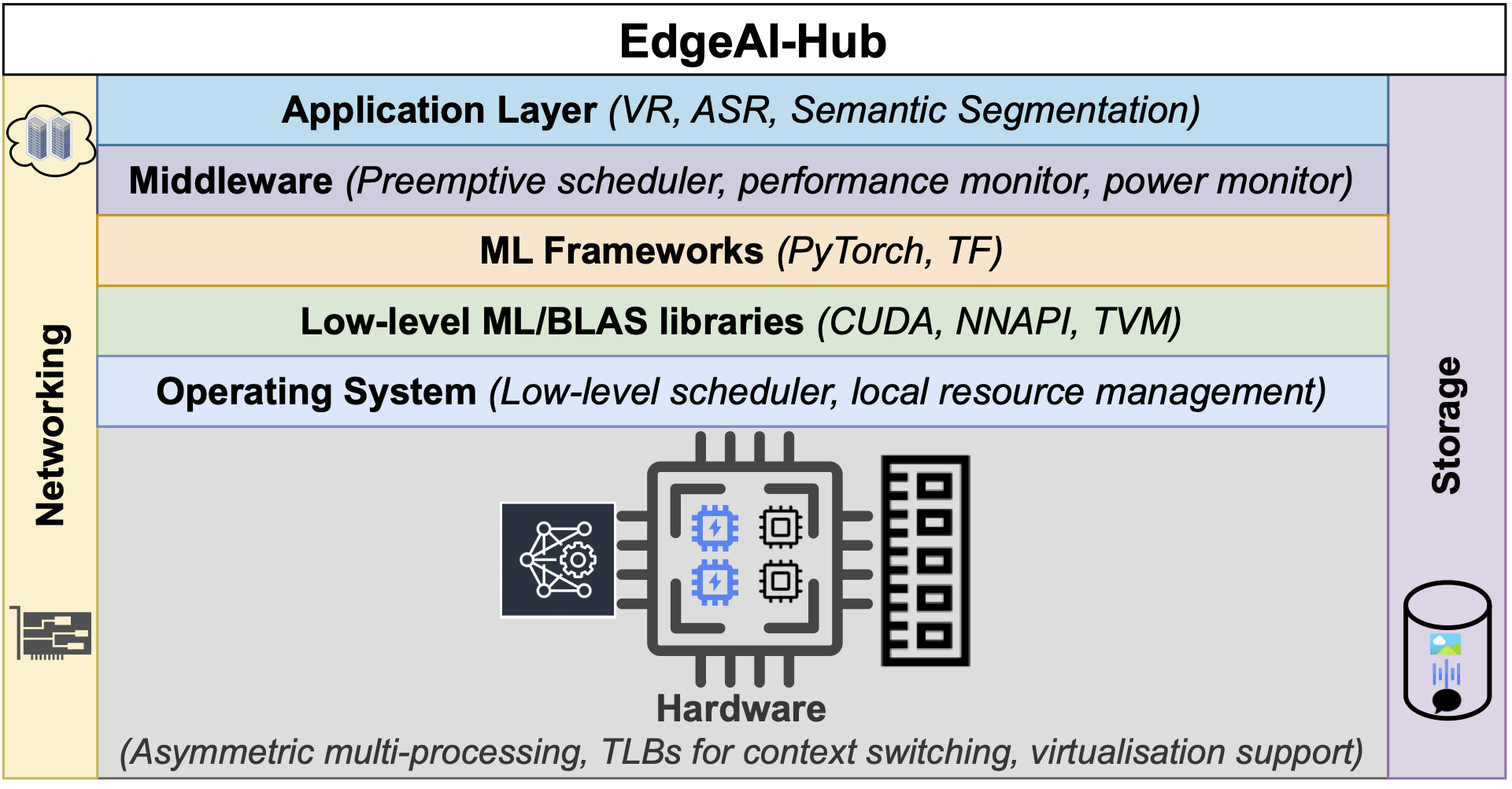}
        \put(-390,4){\footnotesize (a)}
        \put (-190,4) {\footnotesize (b)}
        \label{fig:edgeai_hub}
    }
    \vspace{-0.2cm}
    \caption{Orchestrator reference design (\ref{fig:edgeai_ref}) and EdgeAI-Hub reference stack (\ref{fig:edgeai_hub}).}

\end{figure*}

\subsection{Orchestrator}
{The consumer-edge ecosystem is inherently dynamic, multi-tenant and partially unavailable, mixing mobile and static devices of potentially different owners.
While a peer-to-peer approach may initially seem tempting, we would soon be presented with high preemption rates due to device (un-)availability, along with data inconsistency or model staleness issues. Therefore, we adopt a server-client design where} a central node (orchestrator) is responsible for the subscription and management of resources in the local edge.
The orchestrator should reside in a non-mobile (for availability) device of the edge network (\textit{i.e.}~the EdgeAI-Hub) and should be relatively high-end
in terms of compute and networking capabilities for fine-grained scheduling and high-fidelity cross-channel data communication, respectively. Finally, it can be co-located with an AI-capable device {-- such as a high-end TV or a powerful IoT hub, both of which are persistently available at the edge -- } or can be simply coordinating {them}. {To avoid having a single point of failure or bottlenecks, there can be a secondary orchestrator residing on another capable device if such exists at the edge.}

A reference design of the orchestrator is depicted in Fig.~\ref{fig:edgeai_ref}. Effectively, the orchestrator is the coordinator of the AI-task processes among device resources, {with optional  cloud offloading, where Trusted Execution Environments (TEEs) could ensure privacy}. To this direction, it comprises a \textit{resource manager} to keep track of available resources and dynamic load of devices and a \textit{scheduler} to allocate jobs. Each \textit{device} maintains its own \textit{queue}, with preemptible tasks based on their relative priority. To facilitate resource-to-task matching, a \textit{performance controller} assesses an AI-tasks's runtime on a certain device through analytical or historical estimators.

Upon scheduling a training or inference task, the orchestrator tracks its execution; the respective \textit{controller} monitors the task progress, and handles data and model access between devices or even parameter aggregation, if applicable. This way, context sharing and collaborative learning are enabled in a robust manner, while access to sensitive data remains controlled.

\subsection{EdgeAI-Hub}
Fig.~\ref{fig:edgeai_hub} showcases the technology stack of an EdgeAI-Hub.
At the bottom, we have the hardware that is responsible for running the AI-tasks. The underlying SoC would integrate general-purpose and specialised hardware, \textit{i.e.}~an NPU, to accelerate common DNN operations, from convolutional and fully-connected layers
to Transformer blocks~\cite{fan2022adaptable}. Contrary to smartphone SoCs nowadays, EdgeAI-Hub hardware should be optimised for \textit{i)}~\textit{a broader set of operations}, \textit{ii)}~\textit{multi-precision support}, {\textit{iii)}~\textit{sparsity and dynamic input length support}}, \textit{iv)}~large \textit{multi-level memory} for training with Direct Memory Access (DMA) support for zero-copy distributed ML, \textit{v)}~\textit{virtualisation} for fast context switching and task preemption under multi-user tenancy, and \textit{vi)}~hardware-based \textit{app sandboxing} with TEE~\cite{darknetz} for secure processing of sensitive data.

Network-wise, we envision the EdgeAI-Hub to support simultaneous communication over multiple interfaces, both for supporting different devices as well as for load-balancing and higher throughput. Multiple-Input Multiple-Output (MIMO) over multiple antennas could increase communication capacity, along with mesh networking over proxy repeaters for densely covering the deployment setting. Last, device discovery and handshaking should be made possible over any supported channel, \textit{e.g.}~BLE, NFC, UWB. Extensibility through removable components could also future-proof the EdgeAI-Hub's hardware {and trade energy efficiency for additional bandwidth.}

Data are a first-class citizen in AI and, thus, storage becomes important. We propose a hierarchical storage solution with fast caching for iterative operations, like model training or model/context sharing, and traditional drives for long-term persistence. Hardware-level encryption and user-management with Access Control List (ACL) support from the filesystem is important to ensure security and privacy of data. Finally, redundancy could be offered through hardware (\textit{e.g.}~RAID) or distributed replication.

Moving up, the operating system (OS) is responsible for local resource management and task scheduling. Low-level ML compilers and BLAS libraries would reside on top of the OS, along with the high-level ML framework interpreters for DNN execution.

The middleware and application layers form the top of the stack. The former is responsible for the coordination of distributed execution of tasks, including the orchestrator, preemptive scheduler and performance monitor of local and remote resources. The application may then span across different use-cases from the following section. %

\section{Enabling Upcoming Use-Cases}
\label{sec:use_cases}

There are various existent or upcoming applications that can benefit from the proposed paradigm.

\noindent \textbf{Virtual assistants \& enhanced interaction.}
Home assistants have been progressively penetrating the consumer-edge. However, they largely remain thinly-provisioned, with most computation happening on the cloud. The proliferation of
LLMs that could be -- at least partially -- hosted on premises could significantly improve {privacy}, but also enable new use-cases of
general or context- and user-specific question answering.
Additionally, these models can be combined with other modalities (\textit{e.g.}~vision) and repurposed for more complex tasks in the context of smart homes. %

\noindent \textbf{Virtual spaces.}
AR/VR technologies have been undoubtedly getting more capable {(\textit{e.g.}~Apple Vision Pro)} and the promise of virtual worlds like the Metaverse increasingly relevant. Furthermore, the importance of online communication has been further enhanced by the COVID-19 pandemic and alternative models of collaboration, such as telepresence, have emerged.
As the physical world enters virtual reality, ML plays a central role and new architectures for scene representation or object mapping (\textit{e.g.}~NERF) are utilised, requiring significant computational power {and energy to remain mobile}. Preparing the consumer ecosystem for such workloads {via low-latency multi-device collaborative inference}~\cite{9905603} can be the enabler for such next-gen technologies, be it for multiplayer VR gaming, hybrid collaboration or immersive entertainment.

\noindent {\textbf{Asynchronous collaboration.} Maintaining a work-life balance when working from home can be difficult, especially when collaborating in international and multi-timezone settings. Therefore, tools enabling queryable meeting and document summarisation or multilingual translation (e.g. Microsoft Office co-pilot), along with creative writing or coding assistants (e.g. Github co-pilot) can be valuable tools for asynchronous collaboration. {With privacy being a major business requirement, running such models locally becomes crucial. By locally offloading computation to trusted devices, this can be realistically implemented.}}

\noindent \textbf{Cyber-physical agents.}
Inversely, computer agents have started having physical presence. Robotic agents, in the form of vacuum cleaners, assistants for the elderly or multi-articulated arms for cooking have appeared and push the envelope of what is possible. In this realm, agents may co-exist with one another or with other sensors that can enhance their perception and spatial awareness. Moreover, mobile robots can act as active learning agents for gathering knowledge and reducing model uncertainty.

\noindent \textbf{e-Health.}
{Edge AI has the potential to democratise healthcare by making services available at the point-of-care via automated or assisted diagnostics and treatments~\cite{janbi2022imtidad}, without the need to send patient data to the cloud. Be it through automated diagnostics or treatments, over AI-aided consultations and care,
Edge-AI-driven services can improve efficiency and privacy while lowering risk and cost.
Additionally,
wearables
allow for real-time health and fitness feedback and, coupled with external sensors at home, can provide insights and alerts for health issues in a multi-sensory manner. This can be complemented with digital gastronomy tools~\cite{Han_2020_WACV}, including personalised dietary or cooking instructions, for more complete well-being assistants.}

\section{Challenges}
\label{sec:challenges}

The advent of such new consumer ecosystems brings about several challenges.
These remain largely open issues and we anticipate them becoming prevalent topics of future research.

\subsection{ML-related challenges}

\noindent
{\textbf{Non-IID data, tasks and annotations.} In-the-wild data are highly heterogeneous across clients and may temporally evolve. As such, techniques from domain adaptation, FL and Continual Learning become increasingly relevant to generalisation. Moreover, local data available for training may not be ``clean'' or annotated with labels, thus paving the way for alternative methods, including semi-supervised, unsupervised~\cite{lubana2022orchestra} and active learning~\cite{10471526} for generalisation.}

\noindent
\textbf{Model co-habitation.} Nowadays smart-devices (\textit{e.g.}~smartphones) perform a multitude of tasks concurrently. As such, several DNNs may be cohabitating the device running in parallel. How these are scheduled to be executed efficiently is an open problem \cite{venieris2022multi}.

\noindent
\textbf{Utility \& privacy.} FL is becoming a prevalent paradigm for pushing training to the device, the custodian of data. However, there is a trade-off between global and private utility, as optimisation goals can clash. Adding differential privacy noise to the updates~\cite{mcmahan2017learning}
adds another degree of freedom, trading utility for privacy. Striking a balance between all these factors can be challenging.

\noindent
\textbf{Right to be forgotten.} While learning is the most important task of AI, it is becoming increasingly essential for users to control where their data are being used. As such, model unlearning~\cite{bourtoule2021machine}
becomes a new field of study on how a model can extract and remove knowledge tied to specific data.

\noindent
\textbf{Confidence assessment.} As AI-tasks progressively make their way into the physical world, through robotic assistants, it is necessary to not only make decisions in a binary manner, but also assess the confidence of their decisions before acting, both for safety and interpretability~\cite{kendall2017uncertainties}.

\subsection{System-level challenges}

\noindent
\textbf{System heterogeneity and availability.} Contrary to the datacenter, devices in the wild can be very heterogeneous
and may arbitrarily become unavailable, {due to mobility, power or network connectivity}. As such, they should be treated accordingly, as this can affect the overall performance, robustness or even fairness of the (eco)system.

\noindent
{\textbf{Multi-channel networking.} The network interconnect is a core component in the proposed paradigm, acting as a multi-dimensional bus among devices. This interconnect is largely wireless, potentially lossy and heterogeneous in its implementation and qualitative characteristics. Thus, careful task scheduling and channel load balancing is important for ensuring best QoE.}

\noindent
\textbf{Fault-tolerance \& task preemption.} As resources may become unavailable, high-priority tasks should be given fault-tolerance margins or be restarted on available resources in a transparent-to-the-user manner.

\noindent
\textbf{Interoperability.} Another repercussion to extreme heterogeneity is that devices of different manufacturers, generations and tiers will need to co-operate through a common interconnect. {The same applies to networking interoperability, which can become a bottleneck across generations of devices.} As such, backwards compatibility, standardisation of interfaces and open middleware become requirements to break away from silos and enable device federation.

\noindent
\textbf{Incentives.} An equally important factor for realisation of such a paradigm is incentivisation of the stakeholders: \textit{i)}~Manufacturers to support federated operation of devices and unlocking the potential of old devices; \textit{ii)}~Clients to accept on-premise execution of AI-tasks that can potentially benefit more users (\textit{e.g.}~FL).

\noindent
{\textbf{Market transcendence.} It is unlikely that many consumers will invest on an expensive device ecosystem at once. Therefore, it is {valuable} to consider the consumer market in a stateful manner, where devices pre-exist and come up with a way of gradual transcendence. An example could be firstly co-locating AI-Hubs on premium high-end devices (\textit{e.g.}~TVs) that progressively get marketed as modular components and standalone devices. This also means that connected devices can gradually get cheaper since they do not need to overprovision for bespoke hardware.}

\section{Conclusions}

{We have just started witnessing the revolutionary capabilities of the latest generation of ML models. However, their resource demands are beyond what the current consumer edge can sustain, making them accessible to only a few. In this article, we envision a new {Edge-AI} paradigm, built around EdgeAI-Hubs. These hubs do not only enable complex, resource-intensive applications to run at the consumer edge, but also prioritise user-privacy. To achieve this, we have laid the groundwork with a cross-layer blueprint of this architecture and pinpointed the challenges that lie ahead. Ultimately, our work serves as a foundation to a new, more collaborative model in consumer electronics that can set the stage for how our devices operate in the future.}

\def\refname{REFERENCES}
{\small
\bibliographystyle{plain}
\bibliography{main}

\begin{thebibliography}{10}

\bibitem{codesign}
Mohamed~S. Abdelfattah, Łukasz Dudziak, Thomas Chau, Royson Lee, Hyeji Kim, and Nicholas~D. Lane.
\newblock {Best of Both Worlds: AutoML Codesign of a CNN and its Hardware Accelerator}.
\newblock In {\em Design Automation Conference (DAC)}, 2020.

\bibitem{10471526}
Jin-Hyun Ahn, Yeeun Ma, Seoyun Park, and Cheolwoo You.
\newblock Federated active learning (f-al): An efficient annotation strategy for federated learning.
\newblock {\em IEEE Access}, 12:39261--39269, 2024.

\bibitem{embench2019emdl}
Mario Almeida, Stefanos Laskaridis, Ilias Leontiadis, Stylianos~I. Venieris, and Nicholas~D. Lane.
\newblock {EmBench: Quantifying Performance Variations of Deep Neural Networks across Modern Commodity Devices}.
\newblock In {\em The 3rd International Workshop on Deep Learning for Mobile Systems and Applications (EMDL)}, 2019.

\bibitem{gaugenn2021imc}
Mario Almeida, Stefanos Laskaridis, Abhinav Mehrotra, Lukasz Dudziak, Ilias Leontiadis, and Nicholas~D. Lane.
\newblock {Smart at What Cost? Characterising Mobile Deep Neural Networks in the Wild}.
\newblock In {\em ACM Internet Measurement Conference (IMC)}, 2021.

\bibitem{bommasani2021opportunities}
Rishi Bommasani, Drew~A. Hudson, Ehsan Adeli, et~al.
\newblock On the opportunities and risks of foundation models.
\newblock {\em ArXiv}, 2021.

\bibitem{fl_sys_design2019mlsys}
Keith Bonawitz, Hubert Eichner, Wolfgang Grieskamp, et~al.
\newblock {Towards Federated Learning at Scale: System Design}.
\newblock In {\em Proceedings of Machine Learning and Systems (MLSys)}, 2019.

\bibitem{bonawitz2019federated}
Keith Bonawitz, Fariborz Salehi, Jakub Kone{\v{c}}n{\`y}, Brendan McMahan, and Marco Gruteser.
\newblock {Federated Learning with Autotuned Communication-efficient Secure Aggregation}.
\newblock In {\em 53rd Asilomar Conference on Signals, Systems, and Computers}, 2019.

\bibitem{bourtoule2021machine}
Lucas Bourtoule, Varun Chandrasekaran, Christopher~A Choquette-Choo, Hengrui Jia, Adelin Travers, Baiwu Zhang, David Lie, and Nicolas Papernot.
\newblock {Machine Unlearning}.
\newblock In {\em 2021 IEEE Symposium on Security and Privacy (SP)}, pages 141--159. IEEE, 2021.

\bibitem{gpt3}
Tom Brown, Benjamin Mann, Nick Ryder, Melanie Subbiah, Jared~D Kaplan, Prafulla Dhariwal, Arvind Neelakantan, Pranav Shyam, Girish Sastry, Amanda Askell, et~al.
\newblock Language models are few-shot learners.
\newblock In {\em Advances in Neural Information Processing Systems (NeurIPS)}, 2020.

\bibitem{capc2021iclr}
Christopher~A. Choquette-Choo, Natalie Dullerud, Adam Dziedzic, Yunxiang Zhang, Somesh Jha, Nicolas Papernot, and Xiao Wang.
\newblock {CaPC Learning: Confidential and Private Collaborative Learning}.
\newblock In {\em International Conference on Learning Representations (ICLR)}, 2021.

\bibitem{dang2020should}
Shuping Dang, Osama Amin, Basem Shihada, and Mohamed-Slim Alouini.
\newblock What should 6g be?
\newblock {\em Nature Electronics}, 3(1):20--29, 2020.

\bibitem{desislavov2021compute}
Radosvet Desislavov et~al.
\newblock {Trends in AI inference energy consumption: Beyond the performance-vs-parameter laws of deep learning}.
\newblock {\em Sustainable Computing: Informatics and Systems}, 38:100857, 2023.

\bibitem{fan2022adaptable}
Hongxiang Fan, Thomas Chau, Stylianos~I. Venieris, Royson Lee, Alexandros Kouris, Wayne Luk, Nicholas~D Lane, and Mohamed~S. Abdelfattah.
\newblock {Adaptable Butterfly Accelerator for Attention-based NNs via Hardware and Algorithm Co-design}.
\newblock In {\em International Symposium on Microarchitecture (MICRO)}, 2022.

\bibitem{memory-eff-training-on-device}
In~Gim and JeongGil Ko.
\newblock {Memory-Efficient DNN Training on Mobile Devices}.
\newblock In {\em International Conference on Mobile Systems, Applications and Services (MobiSys)}, 2022.

\bibitem{Han_2020_WACV}
Fangda Han, Ricardo Guerrero, and Vladimir Pavlovic.
\newblock Cookgan: Meal image synthesis from ingredients.
\newblock In {\em Proceedings of the IEEE/CVF Winter Conference on Applications of Computer Vision (WACV)}, March 2020.

\bibitem{fjord}
Samuel Horvath, Stefanos Laskaridis, Mario Almeida, Ilias Leontiadis, Stylianos~I. Venieris, and Nicholas~D. Lane.
\newblock {FjORD: Fair and Accurate Federated Learning under heterogeneous targets with Ordered Dropout}.
\newblock In {\em Advances in Neural Information Processing Systems (NeurIPS)}, 2021.

\bibitem{janbi2022imtidad}
Nourah Janbi et~al.
\newblock Imtidad: A reference architecture and a case study on developing distributed ai services for skin disease diagnosis over cloud, fog and edge.
\newblock {\em Sensors}, 22(5):1854, 2022.

\bibitem{samsung_npu}
Jun-Woo Jang, Sehwan Lee, Dongyoung Kim, Park, et~al.
\newblock {Sparsity-Aware and Re-configurable NPU Architecture for Samsung Flagship Mobile SoC}.
\newblock In {\em International Symposium on Computer Architecture (ISCA)}, 2021.

\bibitem{jiang2018chameleon}
Junchen Jiang, Ganesh Ananthanarayanan, Peter Bodik, Siddhartha Sen, and Ion Stoica.
\newblock {Chameleon: Scalable Adaptation of Video Analytics}.
\newblock In {\em Conference of the ACM Special Interest Group on Data Communication}, 2018.

\bibitem{kendall2017uncertainties}
Alex Kendall and Yarin Gal.
\newblock What uncertainties do we need in bayesian deep learning for computer vision?
\newblock {\em Advances in neural information processing systems}, 30, 2017.

\bibitem{kouris2022multi}
Alexandros Kouris, Stylianos~I Venieris, Stefanos Laskaridis, and Nicholas Lane.
\newblock {Multi-Exit Semantic Segmentation Networks}.
\newblock In {\em ECCV}, 2022.

\bibitem{laskaridis2024melting}
Stefanos Laskaridis, Kleomenis Kateveas, Lorenzo Minto, and Hamed Haddadi.
\newblock Melting point: Mobile evaluation of language transformers.
\newblock {\em arXiv preprint arXiv:2403.12844}, 2024.

\bibitem{10.1145/3469116.3470012}
Stefanos Laskaridis, Alexandros Kouris, and Nicholas~D. Lane.
\newblock {Adaptive Inference through Early-Exit Networks: Design, Challenges and Directions}.
\newblock In {\em International Workshop on Embedded and Mobile Deep Learning (EMDL)}, 2021.

\bibitem{spinn}
Stefanos Laskaridis, Stylianos~I. Venieris, Mario Almeida, Ilias Leontiadis, and Nicholas~D. Lane.
\newblock {SPINN: Synergistic Progressive Inference of Neural Networks over Device and Cloud}.
\newblock In {\em International Conference on Mobile Computing and Networking (MobiCom)}, 2020.

\bibitem{laskaridis2020hapi}
Stefanos Laskaridis, Stylianos~I Venieris, Hyeji Kim, and Nicholas~D Lane.
\newblock {HAPI: Hardware-aware progressive inference}.
\newblock In {\em International Conference on Computer-Aided Design (ICCAD)}, 2020.

\bibitem{lubana2022orchestra}
Ekdeep~Singh Lubana, Chi~Ian Tang, Fahim Kawsar, Robert~P Dick, and Akhil Mathur.
\newblock Orchestra: Unsupervised federated learning via globally consistent clustering.
\newblock {\em arXiv preprint arXiv:2205.11506}, 2022.

\bibitem{9001257}
Peter Mattson, Vijay~Janapa Reddi, Christine Cheng, Cody Coleman, Greg Diamos, David Kanter, Paulius Micikevicius, David Patterson, Guenther Schmuelling, Hanlin Tang, Gu-Yeon Wei, and Carole-Jean Wu.
\newblock {MLPerf: An Industry Standard Benchmark Suite for Machine Learning Performance}.
\newblock {\em IEEE Micro}, 40(2):8--16, 2020.

\bibitem{mcmahan2017learning}
H~Brendan McMahan, Daniel Ramage, Kunal Talwar, and Li~Zhang.
\newblock {Learning Differentially Private Recurrent Language Models}.
\newblock In {\em International Conference on Learning Representations (ICLR)}, 2018.

\bibitem{darknetz}
Fan Mo, Ali~Shahin Shamsabadi, Kleomenis Katevas, Soteris Demetriou, Ilias Leontiadis, Andrea Cavallaro, and Hamed Haddadi.
\newblock {DarkneTZ: Towards Model Privacy at the Edge Using Trusted Execution Environments}.
\newblock In {\em International Conference on Mobile Systems, Applications, and Services (MobiSys)}, 2020.

\bibitem{10.1145/3470974}
Jurn-Gyu Park, Nikil Dutt, and Sung-Soo Lim.
\newblock {An Interpretable Machine Learning Model Enhanced Integrated CPU-GPU DVFS Governor}.
\newblock {\em ACM Trans. Embed. Comput. Syst. (TECS)}, 2021.

\bibitem{Patterson2022-ix}
David Patterson, Joseph Gonzalez, Urs H{\"o}lzle, Quoc Le, Chen Liang, Lluis-Miquel Munguia, Daniel Rothchild, David~R So, Maud Texier, and Jeff Dean.
\newblock {The Carbon Footprint of Machine Learning Training Will Plateau, Then Shrink}.
\newblock {\em Computer}, 2022.

\bibitem{9905603}
Ella Peltonen, Ijaz Ahmad, Atakan Aral, et~al.
\newblock The many faces of edge intelligence.
\newblock {\em IEEE Access}, 10:104769--104782, 2022.

\bibitem{clip}
Alec Radford, Jong~Wook Kim, Chris Hallacy, et~al.
\newblock {Learning Transferable Visual Models From Natural Language Supervision}.
\newblock In {\em International Conference on Machine Learning (ICML)}, 2021.

\bibitem{9586232}
Muhammad Shafique, Theocharis Theocharides, Vijay~Janapa Reddy, and Boris Murmann.
\newblock Tinyml: Current progress, research challenges, and future roadmap.
\newblock In {\em 2021 58th ACM/IEEE Design Automation Conference (DAC)}, pages 1303--1306, 2021.

\bibitem{203326}
Mohammad Shahrad and David Wentzlaff.
\newblock {Towards Deploying Decommissioned Mobile Devices as Cheap Energy-Efficient Compute Nodes}.
\newblock In {\em USENIX Workshop on Hot Topics in Cloud Computing (HotCloud 17)}, 2017.

\bibitem{smith2017federated}
Virginia Smith, Chao-Kai Chiang, Maziar Sanjabi, and Ameet~S Talwalkar.
\newblock {Federated Multi-Task Learning}.
\newblock In {\em Advances in Neural Information Processing Systems (NeurIPS)}, 2017.

\bibitem{multiview}
Surat Teerapittayanon, Bradley McDanel, et~al.
\newblock {Distributed Deep Neural Networks Over the Cloud, the Edge and End Devices}.
\newblock In {\em International Conference on Distributed Computing Systems (ICDCS)}, 2017.

\bibitem{vaswani2017attention}
Ashish Vaswani, Noam Shazeer, Niki Parmar, Jakob Uszkoreit, Llion Jones, Aidan~N Gomez, {\L}ukasz Kaiser, and Illia Polosukhin.
\newblock {Attention is All You Need}.
\newblock In {\em Advances in Neural Information Processing Systems (NeurIPS)}, 2017.

\bibitem{venieris2022multi}
Stylianos~I. Venieris, Christos-Savvas Bouganis, and Nicholas~D. Lane.
\newblock Multiple-deep neural network accelerators for next-generation artificial intelligence systems.
\newblock {\em Computer}, 56(3):70--79, 2023.

\bibitem{wan2024efficient}
Zhongwei Wan, Xin Wang, Che Liu, Samiul Alam, Yu~Zheng, Jiachen Liu, Zhongnan Qu, Shen Yan, Yi~Zhu, Quanlu Zhang, Mosharaf Chowdhury, and Mi~Zhang.
\newblock Efficient large language models: A survey.
\newblock {\em Transactions on Machine Learning Research}, 2024.
\newblock Survey Certification.

\bibitem{wang2019deep}
Erwei Wang, James~J Davis, Ruizhe Zhao, Ho-Cheung Ng, Xinyu Niu, Wayne Luk, Peter~YK Cheung, and George~A Constantinides.
\newblock {Deep Neural Network Approximation for Custom Hardware: Where we've been, where we're going}.
\newblock {\em ACM Computing Surveys (CSUR)}, 52(2):1--39, 2019.

\bibitem{Wu_undated-ns}
Carole-Jean Wu, Ramya Raghavendra, Udit Gupta, et~al.
\newblock {Sustainable AI: Environmental Implications, Challenges and Opportunities}.
\newblock In {\em Proceedings of Machine Learning and Systems (MLSys)}, 2022.

\bibitem{xie2015interference}
Kun Xie, Xin Wang, Xueli Liu, Jigang Wen, and Jiannong Cao.
\newblock {Interference-aware Cooperative Communication in Multi-Radio Multi-Channel Wireless Networks}.
\newblock {\em IEEE Transactions on Computers (TC)}, 2015.

\bibitem{yousefpour2019all}
Ashkan Yousefpour, Caleb Fung, Tam Nguyen, Krishna Kadiyala, Fatemeh Jalali, Amirreza Niakanlahiji, Jian Kong, and Jason~P Jue.
\newblock All one needs to know about fog computing and related edge computing paradigms: A complete survey.
\newblock {\em Journal of Systems Architecture}, 98:289--330, 2019.

\bibitem{fl_homoencryption_usenix}
Chengliang Zhang, Suyi Li, Junzhe Xia, Wei Wang, Feng Yan, and Yang Liu.
\newblock {{BatchCrypt}: Efficient Homomorphic Encryption for {Cross-Silo} Federated Learning}.
\newblock In {\em USENIX Annual Technical Conference (USENIX ATC)}, 2020.

\end{thebibliography}
}

\end{document}